\documentclass[final]{cvpr}

\usepackage{times}
\usepackage{epsfig}
\usepackage{graphicx}
\usepackage{amsmath}
\usepackage{amssymb}
\usepackage{enumitem}
\usepackage{bbm}
\usepackage{tabularx}
\usepackage{booktabs}
\usepackage{varioref}
\usepackage{placeins}

\newcommand{\JW}[1]{\textcolor{green}{}}

\usepackage[pagebackref=true,breaklinks=true,colorlinks,bookmarks=false]{hyperref}

\def\cvprPaperID{10588}  
\def\confYear{CVPR 2021}

\begin{document}

\title{Holistic 3D Human and Scene Mesh Estimation from Single View Images}

        

\author{Zhenzhen Weng, Serena Yeung\\
Stanford University\\
{\tt \{zzweng, syyeung\}@stanford.edu}
}

\maketitle

\begin{abstract}
The 3D world limits the human body pose and the human body pose conveys information about the surrounding objects. Indeed, from a single image of a person placed in an indoor scene, we as humans are adept at resolving ambiguities of the human pose and room layout through our knowledge of the physical laws and prior perception of the plausible object and human poses. However, few computer vision models fully leverage this fact. In this work, we propose a holistically trainable model that perceives the 3D scene from a single RGB image, estimates the camera pose and the room layout, and reconstructs both human body and object meshes. By imposing a set of comprehensive and sophisticated losses on all aspects of the estimations, we show that our model outperforms existing human body mesh methods and indoor scene reconstruction methods. To the best of our knowledge, this is the first model that outputs both object and human predictions at the mesh level, and performs joint optimization on the scene and human poses.
\end{abstract}
\vspace*{-2mm}

\vspace*{-2mm}
\section{Introduction}
Holistic scene perception is key to our human ability to accurately interpret and interact with the 3D world. The human visual system naturally integrates context from actors, objects, and scene layout to infer realistic, robust estimations of the world. Suppose a human is partially included in an image because they are positioned behind a desk. We can still effortlessly extract rich information from the static scene to resolve ambiguities due to the occlusion. Likewise, the appearance of humans also provides useful information about scenes, such as the ground plane and depth of surrounding objects. Humans and objects in scenes jointly manifest spatial occupancies that constrain their relative positions. For computer vision systems to achieve high accuracy in recognizing and interpreting complex scenes, it is therefore important to develop approaches for holistic scene perception and reasoning.

\begin{figure}
\begin{center}
\includegraphics[width=\linewidth]{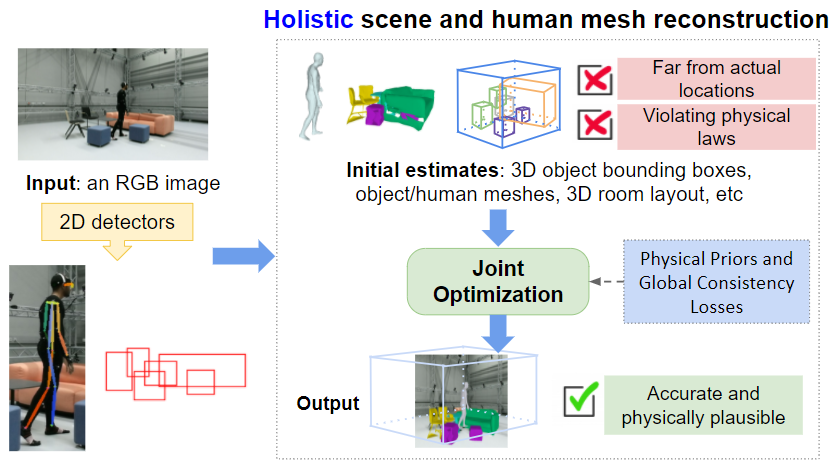}
\end{center}
   \caption{Given a single view RGB image of an indoor scene, our model is able to (i) predict all aspects of the scene (3D object bounding boxes, object and human meshes, 3D room layout, camera pose), and (ii) jointly optimize over a comprehensive set of global consistency losses. The final result is more physically plausible and accurate.
   }
\label{fig:pull_figure}
\end{figure}

In recent years, holistic scene understanding from single view images has gained increasing interest from computer vision researchers. \cite{schwing2013box} \cite{huang2018cooperative} proposed methods for joint reasoning over inanimate scenes, and recovered room layout and 3D object bounding boxes using consistency losses such as a constraint for objects to be enclosed within the room bounding box. \cite{chen2019holistic++} additionally discouraged intersection between object bounding box estimations, and was the first model to bring 3D human pose estimation into the holistic scene understanding problem. It incorporated human-object interaction priors to reason about approximate relations between humans and objects. However all of these works still operate at the relatively coarser level of bounding boxes and joint key points, and are therefore limited in their ability to use precise shapes, surfaces, and physical occupancies to design holistic scene constraints and improve estimation accuracy.

In this work we propose the first single-view, holistic scene understanding method that jointly optimizes over all aspects of 3D human pose, objects, and room layout at the mesh level, to produce state-of-the-art mesh estimations of the scene. Our approach builds on recent advances in mesh prediction. \cite{gao2020tm} \cite{gkioxari2019mesh} \cite{pan2019deep} proposed methods for reconstructing the individual object meshes with varying topological structures. \cite{nie2020total3dunderstanding} builds on \cite{pan2019deep} and proposed the first holistic 3D scene understanding method with mesh reconstruction at the instance level, however they did not consider humans. Recently, \cite{hassan2019resolving} introduced a method for 3D mesh-based human pose estimation, that utilizes physical occupancy information of the static scene to discourage body penetration into the scene. However, \cite{hassan2019resolving} requires the ground truth 3D scans of the scene, and does not perform joint human and scene estimation.

Given a single RGB image, our method simultaneously reconstructs the human body mesh and multiple aspects of the scene – 3D object meshes and bounding boxes, room layout, and camera pose – all in 3D (Figure \ref{fig:pull_figure}). Our approach outputs the SMPL-X (SMPL eXpressive) \cite{pavlakos2019expressive} human mesh model, which fully parameterizes the 3D surface of the human body. It also leverages a variant of the Topology Modification Network (TMN) \cite{gao2020tm}, proposed in \cite{nie2020total3dunderstanding}, as the base model for static object mesh and scene reconstruction. Importantly, we introduce a joint optimization process that incorporates a comprehensive set of physical constraints and priors including 2D/3D reprojection constraints, object-object mesh constraints, object-human mesh constraints, and object/human - room layout constraints, to obtain robust, physically plausible predictions. We perform experimental evaluation on the PiGraphs \cite{savva2016pigraphs} and PROX \cite{hassan2019resolving} datasets and demonstrate that our model outperforms state-of-the-art methods on either 3D scene understanding or 3D human pose estimation.

In summary, our contributions are the following:
\begin{itemize}[noitemsep,topsep=0pt]
\item We propose a holistic trainable model for jointly reconstructing 3D human body meshes and static scene elements (3D object meshes and bounding boxes, room layout, and camera pose) from monocular RGB images. To the best of our knowledge, we are the first to jointly estimate this rich scene understanding at the mesh level.
\item Our  model  does  not  require  any  ground  truth  annotations of the 3D scene or the human poses, and can be directly used on any indoor dataset to produce high quality mesh reconstructions.
\item Through our joint optimization process that incorporates a comprehensive set of physical constraints and priors, we show that our model outperforms prior state-of-the-art methods on either 3D scene understanding or 3D human pose estimation, on the PiGraphs and PROX Quantitative datasets.

\end{itemize}

\section{Related Work}
\textbf{Single View 3D Human Pose Estimation.}
Previous 3D pose estimation methods from single view RGB images can be divided into two types: (i) directly learning 3D human keypoints from 2D image features \cite{sun2018integral}, and (ii) 2D pose estimation with subsequent separate lifting of the 2D coordinates to 3D via deep neural networks \cite{pavllo20193d} \cite{martinez2017simple}. Although these works have showed impressive results on in-the-wild images with relatively clean backgrounds, estimating 3D poses with cluttered background and partial occlusions is still very challenging. Recent works in human body models \cite{loper2015smpl} \cite{pavlakos2019expressive} and single view body mesh reconstruction methods \cite{bogo2016keep} \cite{kanazawa2018end} have pushed the richness of body details available for reasoning, and provide opportunities for bringing novel constraints to the training stage. Recently, \cite{hassan2019resolving} proposed the first 3D human body mesh reconstruction method that takes the static scene into consideration; however they rely on ground truth 3-D scene scans. Our work builds on these directions and is the first to leverage mesh representations of both human and scene in performing holistic estimation of 3D human body and scene meshes jointly. 


\textbf{Holistic Scene Understanding.} The 3D holistic scene understanding problem, in particular 3D scene reconstruction from single view images, has received increasing attention over the past few years. While most of these works have focused on coarser bounding boxes and keypoints as opposed to meshes, methods have differed in model outputs and constraint formulations \cite{huang2018cooperative}\cite{nie2020total3dunderstanding}\cite{chen2019holistic++}. Works such as \cite{huang2018cooperative} have focused on the static scene; \cite{huang2018cooperative} proposed an end-to-end model that learns the 3D room layout, camera pose and 3D object bounding boxes. Drawing insight from the camera projection process and and physical commonsense, \cite{huang2018cooperative} encourages projected 3D bounding boxes to be close to their 2D locations on the image plane, and forces object bounding boxes to be within the room layout bounding box. 

Some works have attempted to incorporate scene/object information in human pose estimation \cite{zanfir2018monocular} \cite{hassan2019resolving} \cite{monszpart2019imapper} \cite{zhang2020perceiving} and/or vice-versa \cite{grabner2011makes}. \cite{zhang2020perceiving} relies on mesh exemplars with annotated contact points, and does not perform full layout/scene reconstruction. \cite{monszpart2019imapper} uses a database of ``scenelets" and works with human skeletons. \cite{hassan2019resolving} utilizes ground truth scene scans. In contrast to these, we consider the more challenging setting of directly estimating scene and human meshes (in general indoor settings), whereas joint mesh estimation is beyond the scope of these works. \cite{chen2019holistic++} jointly tackles two tasks from a single-view image: (i) 3D estimations of object bounding boxes, camera pose, and room layout; and (ii) 3D human keypoints estimation. They used an energy-based inference optimization process that refines direct 3D outputs by jointly reasoning across aspects of the objects and human keypoints. However, their constraint formulations based on 3D bounding boxes and human keypoints are still lacking in precision. Additionally, energy-based models have the disadvantage of an expensive inference step compared to feed-forward models, and \cite{chen2019holistic++}'s MAP estimation method searches over a discrete set of object locations which may give sub-optimal results. In contrast, we impose precise physical constraints at the \emph{mesh level} in our joint optimization procedure and directly back-propagate the underlying neural networks. 


\textbf{Holistic Scene Mesh Reconstruction.}
An emerging line of work attempts to reconstruct richer information about objects in scenes such as depth \cite{shin20193d}, voxel \cite{li2019silhouette} \cite{tulsiani2018factoring}, or mesh representations \cite{gkioxari2019mesh} \cite{nie2020total3dunderstanding}. Meshes contain much richer 3D shape information about the objects, but are generally harder to reconstruct due to the diverse topology of the shapes. Mesh-retrieval methods \cite{hueting2017seethrough} \cite{huang2018holistic} \cite{izadinia2017im2cad} retrieve 3D models from a large 3D model repository, however the size of these repositories remain a bottleneck. Object-wise mesh reconstruction methods \cite{gao2020tm} \cite{wang2018pixel2mesh} \cite{gkioxari2019mesh} \cite{pan2019deep} take a different approach using end-to-end prediction and refinement of the target mesh of individual objects. Recently, \cite{nie2020total3dunderstanding} incorporated an object-wise mesh reconstruction module in their holistic 3D understanding model for static scenes. However, they did not take advantage of the rich information about object shapes that comes with the meshes, and their reconstructed scene meshes are often physically implausible. Although a recent 3D human mesh estimation method \cite{hassan2019resolving} takes advantage of precise object shapes in their constraint formulation, they use ground truth 3D scene scans. In contrast, we estimate \emph{both humans and the static scene} jointly from single view images. 


\begin{figure*}
\begin{center}
\includegraphics[width=0.94\linewidth]{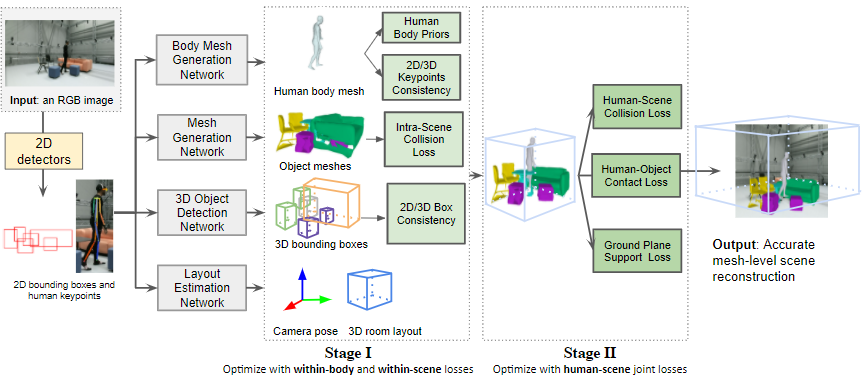}
\end{center}
   \caption{Overview of our model. Given a single RGB image, we first use off-the-shelf 2D detectors to predict the 2D human keypoints and 2D bounding boxes of the objects in the scene. Then, the body mesh network reconstructs a SMPL-X body mesh model through the human keypoints re-projection loss and the human body prior losses. The Mesh Generation Network (MGN) reconstructs the object-wise meshes. 3D Object Detection Network (ODN) predicts the 3D bounding boxes of the objects. Layout Estimation Network (LEN) predicts the camera pose and the 3D room bounding box. In \textbf{Stage I}, the individual modules are optimized with within-body and within-scene losses. In \textbf{Stage II}, the modules fine-tune with the additional human-scene joint losses to achieve consistency and physical plausibility across all aspects of the output.}
\label{fig:system_figure}
\vspace{-3mm}
\end{figure*}

\section{Model}
We introduce a two-stage approach for joint 3D human and scene mesh estimation. In Stage I, we separately parse and reconstruct the human meshes and the 3D scene -- 3D object bounding boxes and meshes, camera pose, and 3D room layout -- to obtain initial estimates. In this stage, holistic reasoning is limited to encouraging physical plausibility within the human only and within the static scene only. Then in Stage II, we jointly minimize global consistency losses across humans and the static scene together, which extends the holistic reasoning to simultaneously improve performance of all sub-tasks.

An overview of our method is illustrated in Figure \ref{fig:system_figure}. In Section \ref{model:representation}, we first define our notation and representation of the 3D scene and our human body mesh model. In Section \ref{model:archi}, we describe the model architectures we use for producing each part of the body and scene estimations. Based on these, in Section \ref{model:losses}, we present our joint optimization process that incorporates a comprehensive set of physical rules and priors --
including reprojection constraints, object-object mesh constraints, object-human mesh constraints, and object/human - room layout constraints -- to perform holistic estimation of both human and scene meshes.

\subsection{Representation}
\label{model:representation}
\textbf{3D Scene.} The input to our model is a 2D image $I \in \mathbb{R}^{(\mathrm{h, w, 3})}$. We use a pre-trained Faster R-CNN \cite{ren2016faster} to obtain initial 2D bounding box estimates $b \in \mathbb{R}^{(4, 2)}$ for each of the $n_{obj}$ objects in the scene. The 2D bounding box centers are represented as $c \in \mathbb{R}^2$. Our representation for the camera pose, room layout, and 3D object bounding boxes and meshes in a scene follows the notation used in \cite{huang2018cooperative}\cite{nie2020total3dunderstanding}. The camera pose is a $3 \times 3$ rotation matrix defined by the pitch and roll angles of the camera system relative to the world system. In the world system, an object bounding box is represented by a 3D bounding box $X \in \mathbb{R}^{(8, 3)}$, which can be determined from its 3D center $C \in \mathbb{R}^3$, spatial size $s \in \mathbb{R}^3$, and orientation angle $\theta \in [-\pi, \pi)$. The cuboid room layout is also represented by a 3D box $X^L \in \mathbb{R}^{(8, 3)}$, and is parameterized in the same manner as an object bounding box. The triangular mesh for object $i$ in the image is represented by its vertices and faces $M_i = (V_i, F_i)$, where $V_i\in \mathbb{R}^{(N_i, 3)}$. $N_i$ is the number of vertices and $F_i$ defines the triangular faces of the mesh. $M_i$ is normalized to fit in a unit cube, and the vertices of the mesh can be converted to the 3D camera coordinate system by translation and rotation as specified by the 3D bounding box parameters.

\textbf{Human Body Model.}
We represent the human body using SMPL-X (SMPL eXpressive) \cite{pavlakos2019expressive}, a generative model that captures how the human body shape varies across a human population, learned from a corpus of registered 3D body, face and hand scans of people of different sizes, genders and nationalities in various poses. SMPL-X extends the SMPL model \cite{loper2015smpl} with fully articulated hands and an expressive face. It is essentially a differentiable function parameterized by shape $\beta_b$, pose $\theta_b$, facial expressions $\psi$ and translation $\gamma$ of the body. The output of SMPL-X is a 3D triangular mesh $M_b = (V_b, F_b)$ that contains $N_b = 10475$ vertices $V_b \in \mathbb{R}^{(N_b, 3)}$ and triangular faces $F_b$.

\subsection{Model Architecture}
\label{model:archi}
\textbf{Body Model.} Since the SMPL-X \cite{pavlakos2019expressive} body model is a fully differentiable function, we simply compute the body loss terms (Section \ref{model:losses}) that are formulated in terms of the vertices and faces of the output human body mesh, and back-propogate the SMPL-X model to find the optimal set of parameters such as the shape and pose of the human body. As in \cite{pavlakos2019expressive}, the parameters of the SMPL-X model are regularized with a set of body priors including a VAE-based body pose prior, and $L_2$ priors on hand pose, facial pose, body shape and facial expressions, penalizing deviation from the neutral state.

\textbf{Scene Models.} 
We use three sub-modules to predict 3D object boxes, camera pose and 3D room layout, and 3D object meshes in the scene, respectively. Specifically, we adopt the Object Detection Network (ODN), Layout Estimation Network (LEN), and Mesh Generation Network (MGN) from \cite{nie2020total3dunderstanding}. For 3D object box prediction, the ODN first takes 2D detections of a Faster R-CNN model trained on LVIS \cite{gupta2019lvis}, extracts appearance features in an object-wise fashion using ResNet-34 \cite{he2016deep}, and encodes the relative position and size between 2D object boxes into geometry features using the method in \cite{hu2018relation}. For each target object, an ``attention sum" is then computed using relational features to other objects \cite{hu2018relation}. Finally, each set of box parameters is regressed using a two-layer MLP. The LEN consists of a ResNet-34 feature extractor and two separate branches with fully-connected layers, one for predicting the camera pose and the other for predicting the 3D room bounding box attributes. Finally, for 3D object mesh prediction, the MGN takes a 2D detection of an object as input and uses ResNet-18 to extract 2D appearance features. Then, the image features concatenated with the one-hot LVIS \cite{gupta2019lvis} object category encoding are fed into the decoder of AtlasNet \cite{groueix2018papier}, which performs mesh deformation from a template sphere mesh. An edge classifier is trained to remove redundant edges from the deformed mesh and a boundary refinement module \cite{pan2019deep} is used to refine the smoothness of boundary edges and output the final mesh. We pre-trained on SUN RGB-D \cite{song2015sun} to initialize the scene models. However, no ground truth annotations are required when training our model on a new dataset.

\vspace{-10pt}
\subsection{Loss Functions and Optimization}
\label{model:losses}
We optimize a comprehensive set of losses based on physically plausible constraints and priors, across two stages of training, to perform holistic estimation of 3D human and scene meshes. These losses can be organized as within-body losses (Stage I), within-scene losses (Stage I), and global human-scene losses (Stage II).


\paragraph{Within-body losses} As part of Stage I of our approach, we first utilize within-body constraints to generate an initial human mesh estimation. Following \cite{hassan2019resolving} \cite{bogo2016keep} \cite{pavlakos2019expressive}, we formulate fitting SMPL-X to monocular images as an optimization problem, and seek to minimize the loss function
\small
\begin{align}
    \mathcal{L}_{\mathrm{body}} =& E(\beta, \theta, \psi, \gamma) \nonumber \\
    =&  E_J+\lambda_{\theta_b}E_{\theta_b} + \lambda_{\theta_f} E_{\theta_f} + \lambda_{\theta_h} E_{\theta_h} + \nonumber \\
    & \lambda_{\mathcal{E}} E_{\mathcal{E}} + \lambda_{\beta} E_{\beta}  + \lambda_{\alpha} E_{\alpha} + \lambda_{\mathcal{P}_{self}} E_{\mathcal{P}_{self}}
\label{eq:1}
\end{align}
\normalsize
Here $E_J$ is the re-projection loss that we use to minimize the weighted robust distance between 2D joints estimated from the RGB image $I$ and the 2D projection of the corresponding 3D joints of SMPL-X. $\theta_b, \theta_f, \theta_h$ are the pose vectors for the body, face (neck, jaw) and the two hands respectively. The terms $E_{\theta_f}$, $E_{\theta_h}$,  $E_{\mathcal{E}}$ and $E_{\beta}$ are $L_2$ priors for the hand pose, facial pose, facial expressions and body shape, penalizing deviations from the neutral state. $E_{\beta}$ is a VAE-based body pose prior called VPoser introduced in \cite{pavlakos2019expressive}. $E_{\alpha}$ is a prior penalizing extreme bending only for elbows and knees. The terms $E_{J}, E_{\theta_b}, E_{\theta_h}, E_{\alpha}, E_{\beta}$ are as described in \cite{pavlakos2019expressive}. $E_{\mathcal{P}_{self}}$ is a penetration penalty for self-penetrations (e.g. hand intersecting knee). The $\lambda$'s are the weights for the terms. 

Our formulation is closest to that in \cite{hassan2019resolving}, which performs human mesh estimation and was built upon \cite{pavlakos2019expressive} with the addition of scene contact ($E_{\mathcal{C}}$) and penetration ($E_{\mathcal{P}}$) terms by assuming access to ground truth scene scans. There are several differences between their full loss function and our formulation in Eq. \ref{eq:1}. First, we do not include any depth related terms, because we wish to perform estimation using solely RGB images whereas \cite{hassan2019resolving} propose model variants leveraging RGB-D information. Second, since we are performing joint estimation of the 3D scene from a monocular RGB image, we are not yet able to reason on scene contact or penetration after only human mesh estimation. So we include only a body self-penetration term in Eq. \ref{eq:1}, which is computed following the approach in \cite{ballan2012motion} \cite{pavlakos2019expressive}\cite{tzionas2016capturing}, and will consider human-scene constraints instead during our global optimization stage.

\vspace{-5pt}
\paragraph{Within-scene losses} 
In Stage I of our approach, we also utilize within-scene constraints to generate an initial static scene estimation. Specifically, we design two within-scene constraints, one for encouraging 2D/3D consistency of the predicted object bounding boxes and the other one for penalizing the collision between the object meshes.

For the first constraint, we utilize the fact that based on the camera projection model, if we project predicted 3D bounding boxes onto the 2D image plane, the projected corners should be close to the 2D bounding box corners. This constraint therefore optimizes both camera pose and 3D bounding boxes. \cite{nie2020total3dunderstanding} imposes a similar loss where they penalize the deviation of the 2D projections of predicted 3D bounding box corners from ground truth 3D bounding box corners for both object bounding boxes and the room bounding box. However, since our model does not rely on any ground truth annotations in our described optimization process, we propose to use our detected 2D bounding boxes as a pseudo ground truth. We show the effectiveness of this loss term in Section \ref{section:experiments}. The formal definition of this term can be written as
\small
\begin{equation}
\mathcal{L}_{\mathrm{scene}}^{J} = \frac{1}{n_{obj}} \sum_{i=1}^{n_{obj}} \mathrm{SmoothL}_1(f(X_i(s_i, C_i, \theta_i)), b_i)
\end{equation}
\normalsize
where $s_i$, $C_i$ $\theta_i$ are the size, centroid and orientation of the object $i$. $b_i$ is the 2D bounding box estimate for object $i$, and $f$ is a differentiable projection function that projects the corners of a 3D bounding box to a 2D image plane. Like \cite{nie2020total3dunderstanding}, we use a smooth $L_1$ loss function comprised of a squared term if the absolute element-wise error falls below a threshold and an $L_1$ term otherwise. 

Our second constraint is a loss term that penalize the collision between reconstructed object meshes. Although some pose estimation works \cite{hassan2019resolving} \cite{jiang2020coherent} have incorporated body collision losses, prior works in scene understanding have not explored this loss, because they either did not have the object shape information necessary to calculate the precise collision \cite{huang2018cooperative} \cite{chen2019holistic++}, or did not take advantage of the object shape information that comes with the meshes \cite{nie2020total3dunderstanding}. We notice that inter-object collision is common in the output of these works. We detect collision using the signed distance field (SDF) of each object. For each object mesh, we voxelize its 3D bounding box into a grid, where for each grid cell center, we calculate its signed distance to the nearest point in the rest of the object meshes in the scene. A negative distance means that this cell center is inside the nearest scene object and denotes penetration. We use a squared sum term of the signed distances of each penetrating grid cell. Formally,
\small
\begin{align}
\mathcal{L}_{\mathrm{scene}}^{\mathcal{P}} = \frac{1}{n_{obj}}\sum_{i=1}^{n_{obj}} \sum_{c_j \in V_{i}}|| d(c_j, M_{-i}) \mathbbm{1}(d(c_j, M_{-i})<0) ||_2^2
\label{eq:2}
\end{align}
\normalsize
where $c_j$ is the center of the $j_{th}$ cell in the voxel grid $V_i$ for object $i$. $d(c_j, M_{-i})$ is the signed distance between the cell center $c_i$ and the scene mesh composed of all object meshes except for object $i$. $\mathbbm{1}$ is an indicator function.

\vspace{-5pt}
\paragraph{Global human-scene losses}
In Stage II of our approach, we jointly fine-tune the human and scene estimation components by imposing additional human-scene losses across the reconstructed human mesh and scene mesh. We consider four types of human-scene losses here.

First, observing that indoor furniture are very likely to be on the floor, we penalize the absolute distance between the object bounding boxes and the ground plane as estimated by the Layout Estimation Network. In the camera coordinate system that we use, $+y$ axis is perpendicular to the ground plane and pointing upward. Hence, we can write this term formally as 
\begin{align}
\mathcal{L}_{\mathrm{joint}}^{\mathrm{obj-ground}} = \frac{1}{n_{obj}}\sum_{i=1}^{n_{\mathrm{obj}}} d(y_{min}(X^L), y_{min}(X_i)))
\end{align}
where $y_{\mathrm{min}}(X)$ returns the minimum $y$ coordinate values of the 3D bounding box $X \in \mathbb{R}^{(8, 3)}$.

Second, like objects in the room, humans need a supporting plane to counteract the gravity. Therefore, we penalize the distance between the lowest point in the human body mesh and the room ground plane. We denote this term as $\mathcal{L}_{\mathrm{joint}}^{\mathrm{body-ground}}$.

Third, we include the contact term $E_{\mathcal{C}}$ from \cite{hassan2019resolving}, although\cite{hassan2019resolving} utlized ground truth scene scans. The intuition is that when humans interact with the scene, they come in contact with it. Thus, \cite{hassan2019resolving} annotates a set of candidate contact vertices $V_C \subset V_b$ across the whole body that come frequently in contact with the world, focusing on the actions of sitting and touching with hands. Formally,
\small
\begin{align}
\mathcal{L}_{\mathrm{joint}}^{\mathcal{C}} & =  \sum_{v_{\mathcal{C}}\in V_{\mathcal{C}}} \rho_{\mathcal{C}} (\min_{v_s \in V_s} || v_{\mathcal{C}}- v_s|| )
\end{align}
\normalsize
where $\rho_{\mathcal{C}}$ denotes a robust Geman-McClure error function \cite{geman1987statistical} for down-weighting vertices in $V_{\mathcal{C}}$ that are far from the nearest vertices the 3D scene mesh $M_s$ which consists of all the meshes in the scene. Note that since we do not have access to (or reconstruct) a floor mesh as in \cite{hassan2019resolving}, we leave out \cite{hassan2019resolving}'s body-floor contact terms; instead, our loss term $\mathcal{L}_{\mathrm{joint}}^{\mathrm{body-ground}}$ encourages contact between the feet and the floor.

Finally, we penalize any collisions between the body mesh and object meshes in the scene. The formulation is similar to Eq. \ref{eq:2}. We call this term $\mathcal{L}_{joint}^{\mathcal{P}}$.

To summarize,our model's total loss is 
\small
\begin{align}
    \mathcal{L}_{\mathrm{total}} &= \mathcal{L}_{\mathrm{body}} + \mathcal{L}_{\mathrm{scene}} + \mathcal{L}_{\mathrm{joint}} \label{eq:loss}
\end{align}
\normalsize
where
\small
\begin{align}
    \mathcal{L}_{\mathrm{scene}} =& \lambda_1 \mathcal{L}_{\mathrm{scene}}^{J} + \lambda_2 \mathcal{L}_{\mathrm{scene}}^{\mathcal{P}} \label{eq:loss_scene} \\
   \mathcal{L}_{\mathrm{joint}} =&  \lambda_{3} \mathcal{L}_{joint}^{\mathrm{obj-ground}} + \lambda_4 \mathcal{L}_{\mathrm{joint}}^{\mathrm{body-ground}}  \nonumber \\
   & + \lambda_{5} \mathcal{L}_{joint}^{\mathcal{C}} + \lambda_{6} \mathcal{L}_{\mathrm{joint}}^{\mathcal{P}}  \label{eq:loss_joint}
\end{align}
\normalsize

In Stage I, only within-body ($\mathcal{L}_{\mathrm{body}}$) and within-scene ($\mathcal{L}_{\mathrm{scene}}$) constraints are used. In Stage II, we add global consistency losses ($ \mathcal{L}_{\mathrm{joint}}$) across humans and the static scene together, and continuously fine-tune the modules to simultaneously improve performance of all sub-tasks. 

\vspace{-8pt}
\section{Experiments}
\label{section:experiments}
\vspace{-2pt}
In this section, we evaluate the performance of our method. Since we are the first to jointly predict and reconstruct both 3D human poses and objects at the mesh level, we compare our model with the state-of-the-art methods for each task. Specifically, we compare with \cite{hassan2019resolving} on human body mesh prediction, \cite{mehta2017vnect}\cite{chen2019holistic++} for 3D human keypoints estimation, and \cite{huang2018cooperative}\cite{chen2019holistic++} for 3D bounding box estimation. 

\subsection{Datasets}
\vspace{-2pt}
\textbf{Pigraphs \cite{savva2016pigraphs}.}
PiGraphs contains $30$ 3D scene scans and $63$ video recordings of five human subjects with skeletal tracking provided by Kinect v2 devices. The dataset contains annotations for 3D human keypoints and 3D object bounding boxes in the scenes. We will perform quantitative evaluation on both of these prediction tasks.

\textbf{PROX Quantitative and Qualitative \cite{hassan2019resolving}.}
PROX Quantitative has $180$ static RGB-D frames and was captured using Vicon and MoSH markers. \cite{hassan2019resolving} placed everyday furniture and objects into the scene to mimic a living room, and performed 3D reconstruction of the scene. The ground truth human body mesh annotations were obtained by placing markers on the body and the fingers, and then using MoSh++ \cite{mahmood2019amass} to convert MoCap data into realistic 3D human meshes represented by a rigged body model. To the best of our knowledge, this is the only available dataset that has both real furniture in a cuboid room as well as a human subject actively interacting with the scene, which makes it ideal for our task. Since PROX Quantitative does not provide ground truth object-level meshes and therefore does not support scene estimation task, we will quantitatively evaluate our model only on the human mesh estimation task. PROX Qualitative \cite{hassan2019resolving} provides $100$K synchronized and spatially calibrated RGB-D recordings of humans in $12$ indoor scenes. While it was released together with PROX Quantitative, it does not have ground truth human mesh annotations. We perform additional qualitative evaluation on this dataset.
\vspace{-4pt}
\subsection{Implementation Details}
\vspace{-2pt}
Given an RGB image of an indoor scene as the input to the model, we first use off-the-shelf 2D detectors to estimate  2D object bounding boxes and 2D human keypoints. For 2D object detections, we use Faster R-CNN \cite{ren2016faster} trained on the LVIS \cite{pavlakos2019expressive} dataset;  
for 2D keypoint detections we use OpenPose \cite{cao2018openpose}. ODN, LEN, and MGN are pretrained on the SUN RGB-D dataset \cite{song2015sun} and Pix3D \cite{sun2018pix3d}, following prior work for our task.

In Stage I, we optimize the SMPL-X body model using only the within-body ($\mathcal{L}_{\mathrm{body}}$) losses. We use L-BFGS optmizer \cite{nocedal2006nonlinear} with learning rate $1\mathrm{e-}3$. For the scene model, we freeze the MGN and the feature extractors components of ODN and LEN, and use Adam \cite{kingma2014adam} optimizer with learning rate $1\mathrm{e-}4$ to back-propagate the linear layers for predicting object bounding box attributes (eg. centroid, orientation), camera pose and 3D room layout. For this part, only the within-scene ($\mathcal{L}_{\mathrm{scene}}$) losses are used.

In Stage II, we add the global consistency losses ($ \mathcal{L}_{\mathrm{joint}}$), and continue fine-tuning of all modules. In this stage, we additionally fix the orientation of the 3D object and room bounding boxes and the camera pose. We train the linear layers for predicting the centroid and the size of the object and room boxes to further refine the 3D location of the objects and the ground plane of the scene. We use the same optimizers as Stage I but with reduced learning rates ($1\mathrm{e-}4$ for L-BFGS \cite{nocedal2006nonlinear} and $5\mathrm{e-}5$ for Adam). 

\subsection{Quantitative Results}
\begin{table}
\begin{center}
\scalebox{0.8}{%
    \begin{tabular}{|l|c|c||l|c|c|}
    \hline
    \multicolumn{3}{|c||}{Object Detection} & \multicolumn{3}{|c|}{Pose Estimation} \\
    \hline
     Methods & 2D IoU & 3D IoU & Methods & 2D (pix) & 3D (m) \\
    \hline\hline
    \cite{huang2018cooperative} &  68.6 & 21.4 & \cite{mehta2017vnect} & 63.9 & 0.732 \\
    \cite{chen2019holistic++}  & 75.1  & 24.9 & \cite{chen2019holistic++} & 15.9  & 0.472 \\
    w/o joint & 74.2 & 25.2 & w/o joint & 15.9 & 0.469  \\
    Ours & \textbf{75.6} & \textbf{26.3}  & Ours & \textbf{15.8}  &  \textbf{0.460} \\
    \hline
    \end{tabular}
}
\end{center}
\vspace{-2mm}
\caption{\textbf{Left}: Quantitative results for 3D scene reconstruction on Pigraphs. Higher IoU values indicate better performance. \textbf{Right}: Quantitative results for human keypoints estimation on Pigraphs. For both 2D (pix) and 3D (m) metrics, lower values are better. ``w/o joint" is the performance of our model without joint optimization.}
\label{table:pigraphs_iou}
\vspace{-4mm}
\end{table}
\vspace{-5pt}

\textbf{3D Object and Human Pose Estimation.}
To show the efficacy of our method in holistic scene understanding, we quantitatively evaluate 3D object detection and 3D human pose estimation on PiGraphs. No prior works for holistic scene understanding have attempted mesh level reconstruction of the scene and human body; both \cite{huang2018cooperative} and \cite{chen2019holistic++} outputs 3D bounding boxes of objects, and \cite{chen2019holistic++} additionally outputs 3D human keypoints. Thus, we evaluate on the same tasks as these baselines. Since our approach is fully based on physical constraints from externally available mesh models, we do not use any of the 3D annotations in PiGraphs for training, as \cite{huang2018cooperative} does. However, we are still able to outperform both (Table \ref{table:pigraphs_iou}), showing the power of leveraging the rich shape information available through meshes.

Following \cite{huang2018cooperative}, for object detection evaluation, we report mean 3D bounding box IoU, as well as 2D IoU between the 2D projections of the 3D object bounding boxes and the ground-truth 2D boxes. For 3D human keypoints evaluation, we extract the $144$ body joints from the fitted SMPL-X model and only keep the ones used in \cite{mehta2017vnect} \cite{chen2019holistic++}, which is a subset of the SMPL-X joints. As in \cite{chen2019holistic++}, we compute the Euclidean distance between the estimated 3D joints and the ground-truth, and average over all joints. For 2D evaluation, we project the estimated 3D keypoints back to the 2D image plane and compute pixel distance to ground truth. 

The quantitative results for both tasks in Table \ref{table:pigraphs_iou} show that our model outperforms both \cite{huang2018cooperative} and \cite{chen2019holistic++} on the 3D object detection task, and \cite{mehta2017vnect} \cite{chen2019holistic++} on the 3D pose estimation task, which illustrates the effectiveness of our method. The boost in 3D performance is significant, because a large source of error of the baseline models come from inaccurate depth estimation of the objects or the humans. Depth estimation from single view images is generally a difficult problem because 2D visual features are limited in suggesting the depth information. We show that the constraints in our joint optimization help to disambiguate the depth information. The improvement on the object bounding box IoUs suggests that applying fine-grained constraints at the mesh level helps with refining coarser details of the objects.

\vspace{-10pt}
\paragraph{Human Mesh Estimation} We quantitatively evaluate our human body mesh estimation results on PROX Quantitative \cite{hassan2019resolving} (Table \ref{table:quant_human}). We follow the evaluation of \cite{hassan2019resolving}, and report the mean per-joint error without/with procrustes alignment (noted as ``PJE" / ``p.PJE"), and the mean vertex-to-vertex error (noted as ``V2V" / ``p.V2V"). Procrustes alignment is a common trick to adjust the predicted 3D vertices for errors in translation, rotation, and scaling. We include the procustes aligned numbers for completion, but note that since our method optimizes all aspects of the human body including translation, rotation and scaling, V2V and PJE are more meaningful quantitative metrics in evaluating the overall quality of the predicted 3D vertices of the mesh.

We compare our body mesh reconstruction method with \cite{hassan2019resolving}, the state-of-the-art human body mesh reconstruction method on PROX Quantitative. \cite{hassan2019resolving} shares the same body loss ($\mathcal{L}_{\mathrm{body}}$) as us; however it imposes contact ($E_{\mathcal{C}}$) and collision ($E_{\mathcal{P}}$) constraints between the human mesh and the ground truth 3D scene scans. In our method, we consider an estimated scene mesh in formulating our losses instead. Therefore, in Table \ref{table:quant_human}, we include quantitative performance of \cite{hassan2019resolving}'s models using ground truth 3D scene scans for reference, and additionally including the following three baselines models for a fair comparison with our model:
\vspace{-1pt}
\begin{itemize}[noitemsep,topsep=0pt]
    \item \cite{hassan2019resolving} (body terms only): \cite{hassan2019resolving}, without using scene terms (since these utilize a ground truth scene).
    \item \cite{hassan2019resolving} + estimated scene: \cite{hassan2019resolving} with their contact ($E_{\mathcal{C}}$) and collision ($E_{\mathcal{P}}$) terms calculated using the 3D scene mesh predicted by \cite{nie2020total3dunderstanding} (our base scene model).
    \item \cite{hassan2019resolving} + w/in-scene losses: \cite{hassan2019resolving} with their contact ($E_{\mathcal{C}}$) and collision ($E_{\mathcal{P}}$) terms calculated using an optimized scene mesh (the base scene model \cite{nie2020total3dunderstanding}, plus our within-scene losses $\mathcal{L}_{\mathrm{scene}}$). 
\end{itemize}
\vspace{-1pt}
Our model outperforms all three baselines that do not use ground truth scene scans (bottom half of Table \ref{table:quant_human}), and is competitive to \cite{hassan2019resolving}'s models using ground truth scene scans (top half). This shows the effectiveness of our scene mesh estimation in refining the human meshes, and that simply adding estimated scenes to \cite{hassan2019resolving} is not sufficient. The gap between \cite{hassan2019resolving} + w/in-scene losses and Ours highlights the utility of our joint optimization process.

\begin{table}
\begin{center}
\scalebox{0.9}{%
    \begin{tabular}{|l|c|c||c|c|}
    \hline
     \multicolumn{5}{|c|}{\textbf{with} ground truth 3D scene scans} \\
    \hline\hline
        & V2V & PJE & p.V2V & p.PJE \\
    \hline\hline
    \cite{hassan2019resolving} (including $E_{\mathcal{C}}$) &  208.03 & 208.57 & 72.76 & 60.95\\
    \cite{hassan2019resolving} (including $E_{\mathcal{P}}$) &  190.07 & 190.38 & 73.73 & 62.38\\
    Full \cite{hassan2019resolving} ($E_{\mathcal{C}}$ + $E_{\mathcal{P}}$) &  167.08 & 166.51 & 71.97 & 61.14 \\
    \hline \hline 
    \multicolumn{5}{|c|}{\textbf{without} ground truth 3D scene scans} \\
    \hline\hline 
    \cite{hassan2019resolving} (body terms only)  &  220.27 & 218.06 & 73.24 & \textbf{60.80}\\
    \cite{hassan2019resolving} + estimated scene & 224.53 & 220.47& 73.49& 61.32\\
    \cite{hassan2019resolving} + w/in-scene losses & 212.48 & 209.67 & 73.13& 62.06\\
    Ours & \textbf{192.21} & \textbf{190.78} & \textbf{72.72} & 61.01 \\
    \hline
    \end{tabular}
}
\end{center}
\vspace{-2mm}
\caption{Quantitative results for human mesh estimation on PROX Quantitative. Top half of the table contains the performance of \cite{hassan2019resolving}'s models that use ground truth 3D scene scans in optimizing the human body model. Bottom half of the table contains the baseline models that are most comparable to ours, because no ground truth 3D scene scans are used during training. We highlight the best numbers among the models that do not require ground truth scene scans.}
\label{table:quant_human}
\end{table}

\begin{figure*}[t]
\begin{center}
   \includegraphics[width=0.93\textwidth]{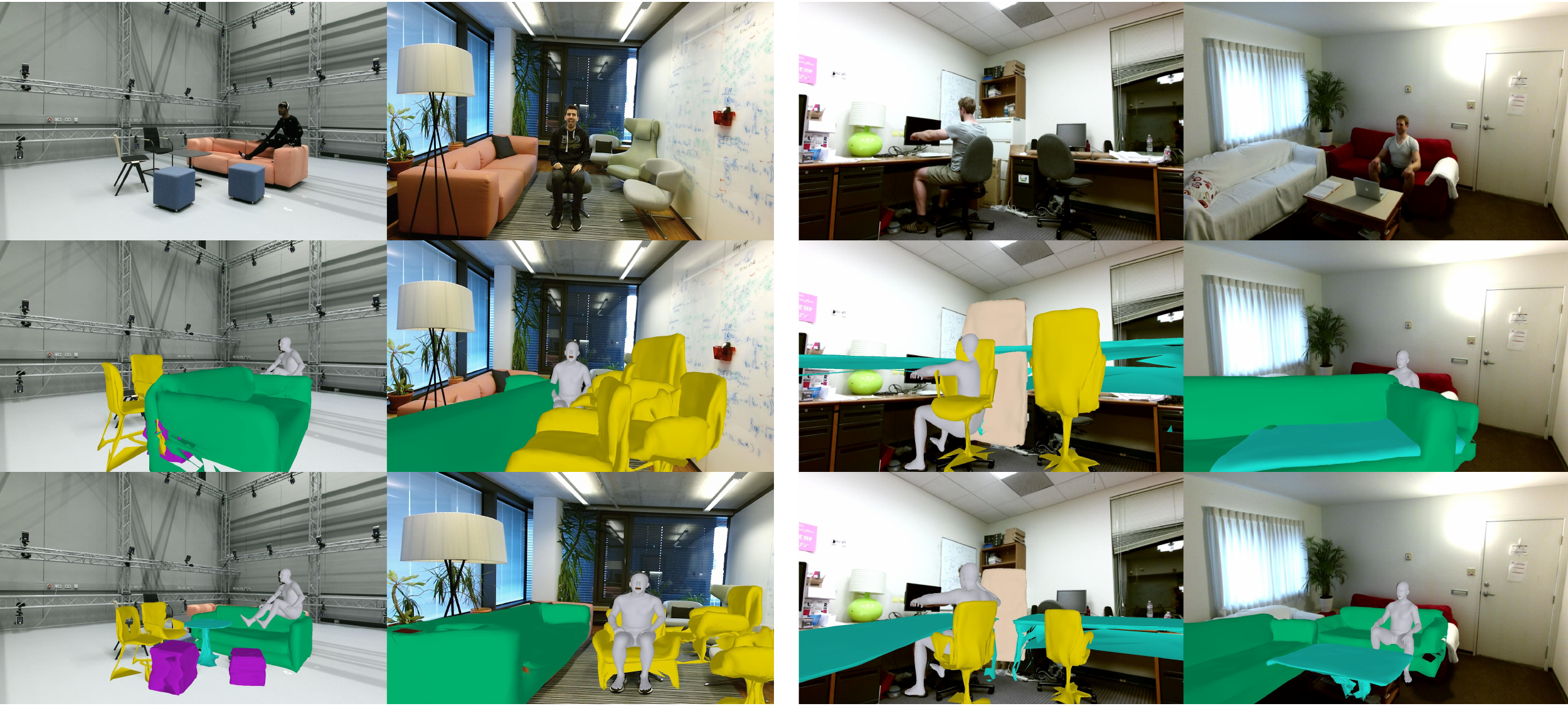}
\end{center}
\vspace{-2mm}
   \caption{\textbf{Left half}: Qualitative results on PROX Quantitative and Qualitative datasets. The left frames is from PROX Quantitative. The right frame is from PROX Qualitative.
   \textbf{Right half}: Qualitative results on Pigraphs dataset. From top to bottom are the RGB input, the direct output from the scene and body mesh without any optimization, and the final mesh with the joint optimization.
   }
\label{fig:quali_prox}
\vspace{-2mm}
\end{figure*}
\subsection{Ablation Analysis}
To analyze the contributions of different losses, we compare variants of our proposed full model. In Tables \ref{ablated:human} and \ref{ablated:box}, we compare quantitative results on the human body mesh prediction and 3D object detection tasks as we take out each one of the losses in Eqs. \ref{eq:loss_scene} and \ref{eq:loss_joint}, except for the essential body loss ($\mathcal{L}_{\mathrm{body}}$) and box re-projection loss ($\mathcal{L}_{\mathrm{scene}}^J$). We observe that all of the losses are essential in improving both the scene estimation and body estimation tasks. The joint losses $\mathcal{L}_{\mathrm{joint}}^{\mathrm{object-ground}},  \mathcal{L}_{\mathrm{joint}}^{\mathcal{C}}$, and $\mathcal{L}_{\mathrm{joint}}^{\mathcal{P}}$ play an essential role in jointly improving the global consistency, which boosts the performance of human body mesh reconstruction task. In particular, $\mathcal{L}_{\mathrm{joint}}^{\mathcal{P}}$ seems to be the most important term in refining the body meshes. The $\mathcal{L}_{\mathrm{joint}}^{\mathrm{object-ground}}$ and $\mathcal{L}_{\mathrm{joint}}^{\mathrm{body-ground}}$ terms improves the ground plane estimation, which helps the 3D object detection task significantly.
\vspace*{-2mm}
\subsection{Qualitative Results}
Figure \ref{fig:quali_prox} shows qualitative results of our models on the PROX Quantitative and Qualitative, and PiGraphs datasets. We observe that the direct output of the scene model (pretrained on SUN-RGBD and Pix3D) without our holistic optimization contains inaccurate object attributes. Our proposed joint optimization method improves the overall accuracy of the predictions by constraining the orientations, positions and the sizes of the objects to be realistic with respect to each other. Also, human pose estimation task helps the optimization of the scene - the chair that the human sits on tend to have more accurate orientations than the other two chairs (column 2).
Besides, the initially estimated ground plane could be very inaccurate (column 3), and our joint optimization process helps adjust the ground plane and improve the location of all objects at the same time. Although not obvious from the qualitative results in Figure \ref{fig:quali_prox}, the estimated scene mesh helps refining the 3D locations of the human body mesh vertices through the joint losses, which is supported by our quantitative results in Tables \ref{table:quant_human} and \ref{ablated:human}. Finally, we show additional qualitative results in Section 2 of the Supplementary, and we discuss limitations and failure cases in Section 3 of the Supplementary.

\begin{table}
\begin{center}
\small
\scalebox{0.92}{%
    \begin{tabular}{|l||c|c||c|c|}
    \hline
       Metrics & V2V & PJE & p. V2V  & p. PJE\\
    \hline\hline
    w/o $\mathcal{L}_{\mathrm{scene}}^{\mathcal{P}}$ & 200.43 & 194.27 & 73.20 & 62.76 \\
    w/o $\mathcal{L}_{\mathrm{joint}}^{\mathrm{body-ground}}$  & 192.18 & 190.84 & 72.21 & 62.39 \\
    w/o $\mathcal{L}_{\mathrm{joint}}^{\mathrm{object-ground}}$  & 196.32 & 193.43 & 72.47 & 62.00\\
    w/o $\mathcal{L}_{\mathrm{joint}}^{\mathcal{C}}$ & 196.48 & 194.32 & 73.24 & 62.96\\
    w/o $\mathcal{L}_{\mathrm{joint}}^{\mathcal{P}}$  &  212.24 & 213.26 & 73.64 & 62.90 \\
    Full model & 192.21 & 190.78 & 72.72 & 61.01 \\
    \hline
    \end{tabular}   
}
\normalsize
\end{center}
\vspace{-2mm}
\caption{Ablations for human mesh estimation on PROX Quant.}
\label{ablated:human}
\vspace{-2mm} 
\end{table}

\begin{table}
\begin{center}
\small
\scalebox{0.92}{%
    \begin{tabular}{|l||c|c||c|c|}
    \hline
       Tasks & \multicolumn{2}{c||}{Object Detection} & \multicolumn{2}{c|}{Pose Estimation} \\
    \hline \hline
     Metrics    & IoU$_{2D}$  & IoU$_{3D}$ & 2D (pix) & 3D (m)\\
    \hline
    w/o $\mathcal{L}_{\mathrm{scene}}^{\mathcal{P}}$ & 58.1 & 19.1 & 16.5 & 0.472 \\
    w/o $\mathcal{L}_{\mathrm{joint}}^{\mathrm{body-grnd}}$ & 52.6 & 10.3 & 16.3 & 0.463 \\
    w/o $\mathcal{L}_{\mathrm{joint}}^{\mathrm{obj.-grnd}}$ & 49.3 & 11.2 & 17.9 & 0.523  \\
    w/o $\mathcal{L}_{\mathrm{joint}}^{\mathcal{C}}$ & 74.6 & 26.4 & 18.4 & 0.493 \\
    w/o $\mathcal{L}_{\mathrm{joint}}^{\mathcal{P}}$ & 73.2 & 24.7 & 21.6 & 0.540 \\
    Full model & 75.6 & 26.3 & 15.8  &  0.460\\
    \hline
    \end{tabular}
}
\normalsize
\end{center}
\vspace{-2mm}
\caption{Ablation results on PiGraphs. For the IoU metrics, higher values indicate better performance. For the pose estimation metrics (2D (pix) and 3D (m)), lower values are better.}
\label{ablated:box}
\vspace{-3mm}
\end{table}

\vspace*{-1mm}
\section{Conclusion}
In this work, we focus on the challenging problem of single view holistic reconstruction and joint optimization of human pose together with static scene. We propose the first holistically trainable model for reconstructing and jointly estimating both 3D human pose and 3D scene at the mesh level. Through a joint optimization process that incorporates a comprehensive set of physical plausibility and priors, we show that our model outperforms state-of-the-art methods on either 3D scene understanding or 3D human pose estimation, on the PiGraphs and PROX Quantitative datasets.
\vspace*{-2mm}
\paragraph{Acknowledgements} This material is based upon work supported by the National Science Foundation under Grant No. 2026498, as well as a seed grant from the Institute for Human-Centered Artificial Intelligence (HAI) at Stanford University.
\section*{Appendix}

\subsection*{A. Additional Training Details}
We direct the readers to \cite{nietotal3dunderstanding} for camera/world system setting and details on the network architecture of ODN, LEN and MGN. Here we elaborate on the training details.
\paragraph{Stage I}
In Stage I, we optimize the SMPL-X body model using only the within-body ($\mathcal{L}_{\mathrm{body}}$) losses. We instantiate a body model for each human in the frame, and use L-BFGS optmizer \cite{nocedal2006nonlinear} with learning rate $1e-3$ to learn the optimal body parameters (e.g. body shape, pose, translation). First, the translation vector of the body model is optimized for $20$ iterations with only the human keypoints re-projection loss. This step is used to roughly position the body model in the camera coordinate system. Then, all the within-body loss terms are considered and the entire body model is optimized for $80$ iterations.

For the scene model, we freeze the MGN and the feature extractors components of ODN and LEN, and use Adam \cite{kingma2014adam} optimizer with learning rate $1e-4$ with weight decay $1e-4$ to back-propagate the linear layers for predicting object bounding box attributes (eg. centroid, orientation), camera pose and 3D room layout. For this part, only the within-scene ($\mathcal{L}_{\mathrm{scene}}$) losses are used. For each frame, the scene model is optimized for $150$ iterations. 

\paragraph{Stage II}
In Stage II, we add the global consistency losses ($\mathcal{L}_{\mathrm{joint}}$), and continue fine-tuning of all modules. In this stage, we additionally fix the orientation of the 3D object and room bounding boxes and the camera pose. We train the linear layers for predicting the centroid and the size of the object and room boxes to further refine the 3D location of the objects and the ground plane of the scene. We use the same optimizers as Stage I but with reduced learning rates ($1e-4$ for L-BFGS and $5e-5$ for Adam). The body model and scene model are optimized alternately for $20$ iterations. The hyperparamters used are $\lambda_1=1$, $\lambda_2=0.1$, $\lambda_3=10$, $\lambda_4=20$, $\lambda_5=1e3$, $\lambda_6=1e2$.

\subsection*{B. Additional Qualitative Results}
\begin{figure*}
\centering
\includegraphics[width=\textwidth]{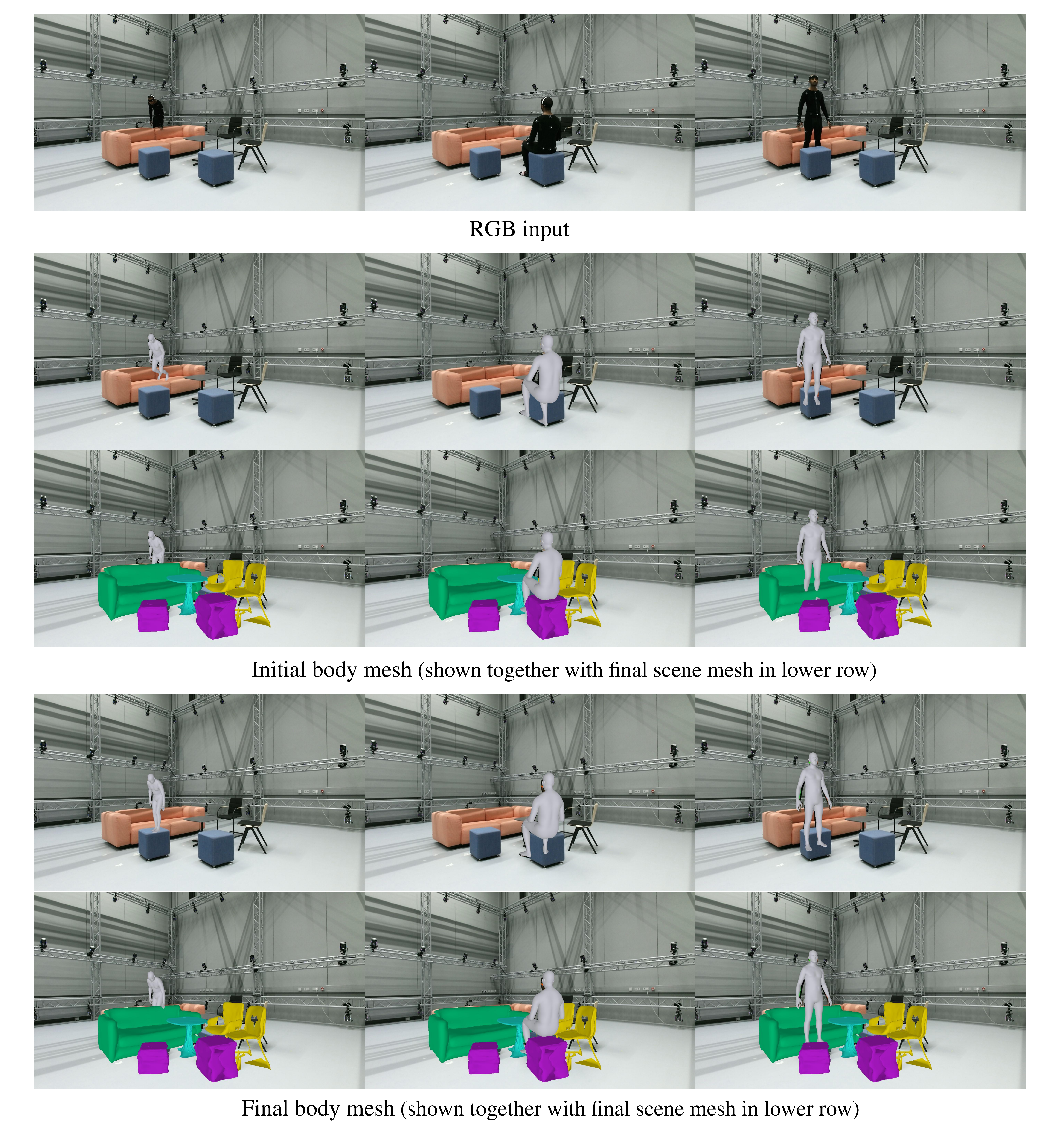}
\caption{Qualitative Results on PROX Quantitative. Each column shows the result on one frame from the PROX Quantitative recordings. From top to bottom are: (1) the RGB input, (2) the initial body mesh (shown together with the final scene mesh in lower row), (3) the final body mesh (shown together with the final scene mesh in lower row).
}
\label{fig:quali}
\end{figure*}

In Figure \ref{fig:quali} we show qualitative examples in PROX Quantitative \cite{hassan2019resolving} where the scene estimation task significantly helps the body estimation task. These are complementary examples to those in the main paper, which showed that the human body estimation task helps the scene estimation.

From Figure \ref{fig:quali} we can see that the initial body meshes are either not physically plausible (column $1$), or are intersecting with the scene (column $2$, $3$). Using the human-scene joint optimization method proposed in our paper, the final body meshes are much more realistic. Note that since we are overlaying the meshes on the 2D images, we can still see the legs behind the furniture after the joint optimization. However, there is no mesh intersection in the 3D coordinate system. 

\begin{figure*}
\centering
\includegraphics[width=1.1\textwidth]{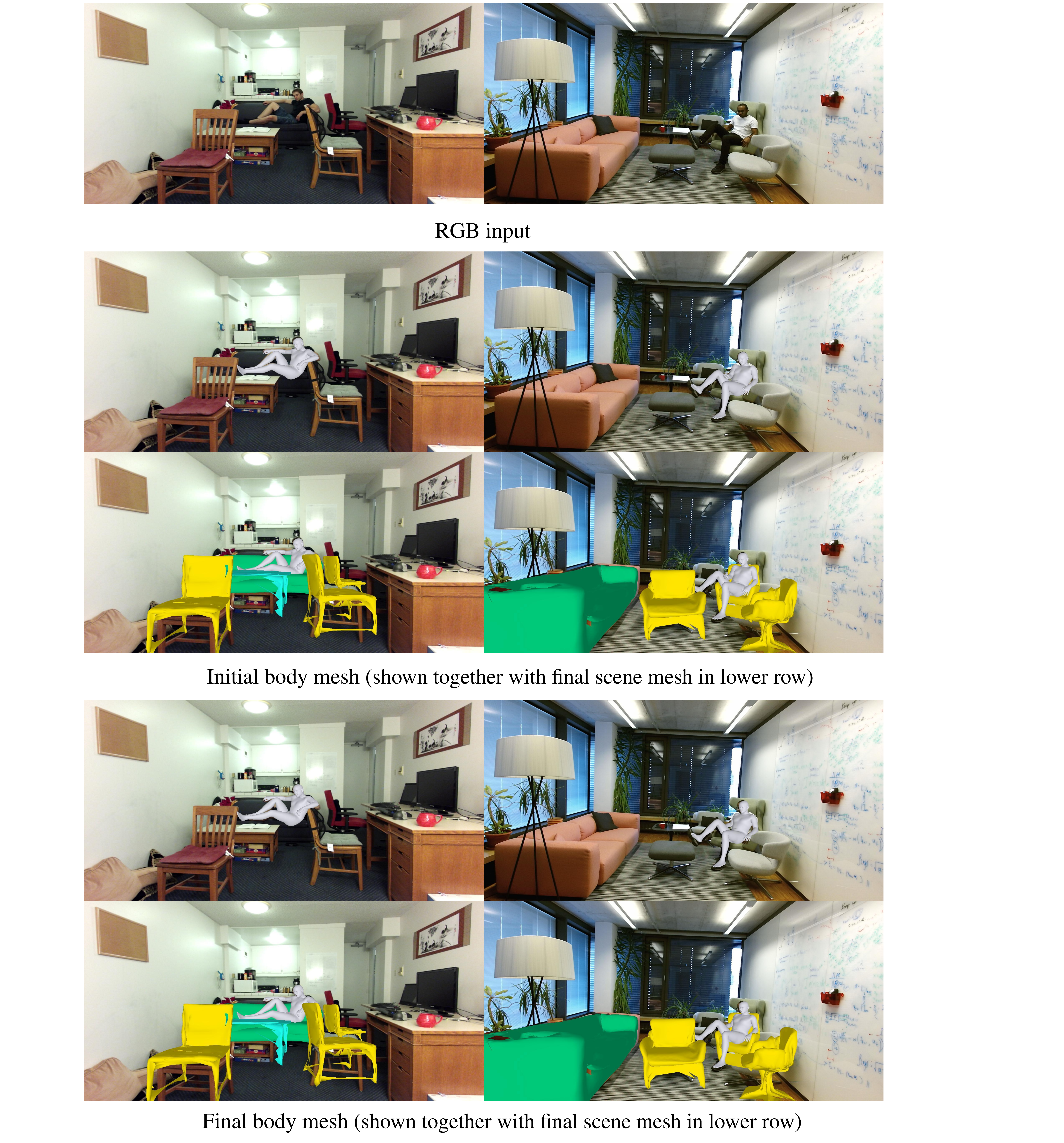}
\caption{Left: Qualitative example on PiGraphs. Right: Qualitative example on PROX Qualitative.  From top to bottom are (1) the RGB input, (2) the initial body mesh (shown together with the final scene mesh in lower row), (3) the final body mesh (shown together with the final scene mesh in lower row).
}
\label{fig:quali1}
\end{figure*}

In Figure \ref{fig:quali1} we show similar results on PiGraphs and PROX Qualitative. We show that the final body meshes in both examples improve through the human-scene optimization stage. In the PiGraphs example, the human body is lifted to reduce the intersection with the sofa. In the PROX Qualitative example, the right hand of the human is occluded so the 2D keypoints predicted by OpenPose \cite{cao2018openpose} do not include the keypoints on the right hand. As a result, the initial hand pose is far from the ground truth. However, through the human-scene optimization that encourages contact between the scene and the hands, the hand pose ended up closer to ground truth.

\begin{figure*}[t]
\begin{center}
   \includegraphics[width=0.95\linewidth]{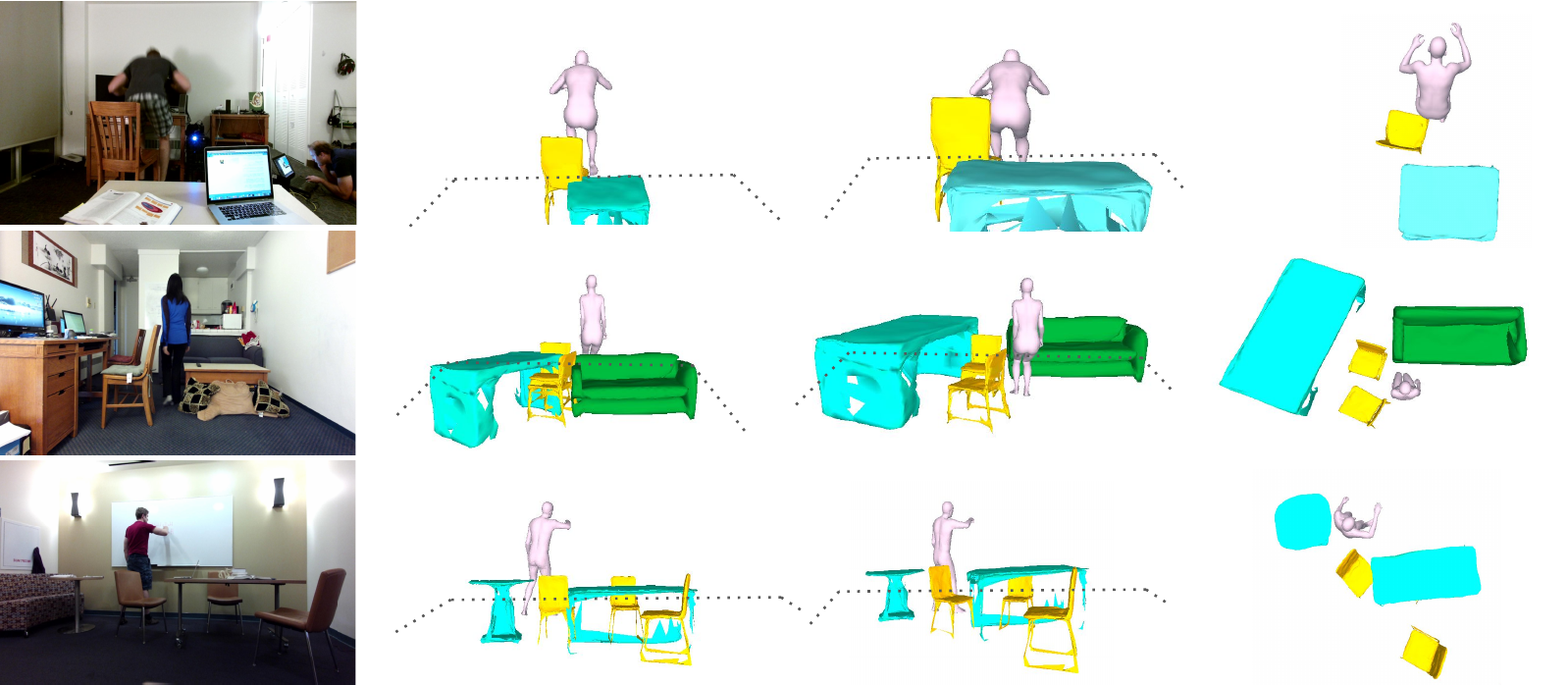}
\end{center}
\caption{Additional qualitative results with two viewpoints. From left to right are: RGB input, scene w/o optimization, scene w/ optimization (front view and top view).}
\label{fig:rebuttal}
\end{figure*}

\subsection*{C. Limitations and Failure Cases}
Our method is limited by the performance of the 2D detectors and the mesh generation network. Examples are in rows 1 (missing desk) and 2 (missing coffee table) of Figure \ref{fig:rebuttal}. Since during joint optimization, the
base object mesh structures are not altered, the mesh generation network decides the quality of the generated meshes. Another failure case is due to difficulty or lack of useful physical hints from the scene. When objects and humans are sparsely allocated, the designed losses are not helpful in adjusting their positions. For instance, Figure \ref{fig:rebuttal}, row 2 shows the incorrect orientation of the chair in the back. In the right columns of Figure 3 in the paper, where the scenes have more occlusion, the ground plane estimation has a small shift away from the actual ground plane.

\newpage

{\small
\bibliographystyle{ieee_fullname}
\bibliography{egbib}
}


\end{document}